\documentclass[twoside,11pt]{article}

\usepackage{jmlr2e}
\usepackage{amsmath}
\usepackage{algorithm}
\usepackage{algorithmic}
\usepackage{subfigure}
\usepackage{graphicx}
\usepackage{bbm}
\usepackage{tikz}
\usetikzlibrary{arrows}
\usetikzlibrary{graphs}

\newcommand{\mb}{\mathbf}
\newcommand{\mc}{\mathcal}
\newcommand{\direct}{\textsc{Direct}}
\newcommand{\lipo}{\textsc{LIPO}}

\newcommand{\mcs}{\textsc{MCS}}
\newcommand{\pso}{\textsc{PSO}}
\newcommand{\gadam}{\textsc{Gadam}}
\newcommand{\sce}{\textsc{SCE}}
\newcommand{\de}{\textsc{DE}}
\newcommand{\ga}{\textsc{GA}}
\newcommand{\es}{\textsc{ES}}
\newcommand{\cmaes}{\textsc{CMA-ES}}
\newcommand{\bs}{\boldsymbol}

\ShortHeadings{IFM LAB TUTORIAL SERIES \# 3, COPYRIGHT \copyright IFM LAB}{JIAWEI ZHANG, IFM LAB DIRECTOR}
\firstpageno{1}

\begin{document}

\title{Derivative-Free Global Optimization Algorithms:\\ Population based Methods and Random Search Approaches}

\author{\name Jiawei Zhang \email jiawei@ifmlab.org \\
	\addr{Founder and Director}\\
       {Information Fusion and Mining Laboratory}\\
       (First Version: March 2019; Revision: April 2019.)}

\maketitle

\begin{abstract}

In this paper, we will provide an introduction to the derivative-free optimization algorithms which can be potentially applied to train deep learning models. Existing deep learning model training is mostly based on the back propagation algorithm, which updates the model variables layers by layers with the gradient descent algorithm or its variants. However, the objective functions of deep learning models to be optimized are usually non-convex and the gradient descent algorithms based on the first-order derivative can get stuck into the local optima very easily. To resolve such a problem, various local or global optimization algorithms have been proposed, which can help improve the training of deep learning models greatly. The representative examples include the Bayesian methods, Shubert-Piyavskii algorithm, {\direct}, {\lipo}, {\mcs}, {\ga}, {\sce}, {\de}, {\pso}, {\es}, {\cmaes}, hill climbing and simulated annealing, etc. This is a follow-up paper of \cite{zhang2019derivativefree_part1}, and we will introduce the population based optimization algorithms, e.g., {\ga}, {\sce}, {\de}, {\pso}, {\es} and {\cmaes}, and random search algorithms, e.g., hill climbing and simulated annealing, in this paper. For the introduction to the other derivative-free optimization algorithms, please refer to \cite{zhang2019derivativefree_part1} for more information.

\end{abstract}


\tableofcontents

\section{Introduction}\label{sec:intro}

This is a follow-up paper of \cite{zhang2019derivativefree_part1}. To make it self-contained, we will briefly introduce the learning settings again as follows. The training set for optimizing the deep learning models can be represented as $\mc{T} = \{(\mb{x}_1, \mb{y}_1), (\mb{x}_2, \mb{y}_2), \cdots, (\mb{x}_n, \mb{y}_n)\}$, which involves $n$ pairs of feature-label instances. Formally, for each data instance, its feature vector $\mb{x}_i \in \mathbb{R}^{d_x}, \forall i \in \{1, 2, \cdots, n\}$ and label vector $\mb{y}_i \in \mathbb{R}^{d_y}, \forall i \in \{1, 2, \cdots, n\}$ are of dimensions $d_x$ and $d_y$ respectively. The deep learning models define a mapping $F(\cdot; \boldsymbol{\theta}): \mc{X} \to \mc{Y}$, which projects the data instances from the feature space $\mc{X}$ to the label space $\mc{Y}$. In the above representation of function $F(\cdot, \bs{\theta})$, vector $\boldsymbol{\theta} \in \Theta$ contains the variables involved in the deep learning model and $\Theta$ denotes the variable inference space. Formally, we can denote the dimension of variable vector $\bs{\theta}$ as $d_{\theta}$, which will be used when introducing the algorithms later. Given one data instance featured by vector $\mb{x}_i \in \mc{X}$, we can denote its prediction label vector by the deep learning model as $\hat{\mb{y}}_i = F(\mb{x}_i; \boldsymbol{\theta})$. Compared against its true label vector $\mb{y}_i$, we can denote the introduced loss for instance $\mb{x}_i$ as $\ell (\hat{\mb{y}}_i, \mb{y}_i)$. Several frequently used loss representations have been introduced in \cite{zhang2019gradient}, and we will not redefine them here again. For all the data instances in the training set, we can represent the total loss term as 
\begin{equation}
\mc{L}(\boldsymbol{\theta}) = \mc{L}(\boldsymbol{\theta}; \mc{T}) = \sum_{(\mb{x}_i, \mb{y}_i) \in \mc{T}} \ell (\hat{\mb{y}}_i, \mb{y}_i).
\end{equation}
And the deep model learning can be formally denoted as the following function:
\begin{equation}
\min_{\boldsymbol{\theta} \in \Theta} \mc{L}(\boldsymbol{\theta}),
\end{equation}
which is also the main objective function to be studied in this paper.

We need to add a remark here, the above objective function defines a minimization problem. Meanwhile, when introducing some of the optimization algorithms in the following sections, we may assume the objective function to be a maximization function instead for simplicity. The above objective can be transformed into a maximization problem easily by introducing a new term $\mc{L}'(\boldsymbol{\theta}) = -\mc{L}(\boldsymbol{\theta})$. We will clearly indicate it when the algorithm is introduced for a maximization problem.

In the following part of this paper, we will introduce the derivative-free optimization algorithms that can be potentially used to resolve the above objective function. To be more specific, this paper covers the introduction to the population based algorithms (e.g., {\ga}, {\sce}, {\de}, {\pso}, {\es} and {\cmaes}) and the random search based optimization algorithms (e.g., hill climbing and simulated annealing). If the readers are interested in other derivative-free optimization algorithms, you can refer to our previous article \cite{zhang2019derivativefree_part1} for more information.

%----------------------------------------------------------------------------------------------
%----------------------------------------------------------------------------------------------
%----------------------------------------------------------------------------------------------

\section{Population based Algorithm for Global Optimization}\label{sec:ea}

In this section, we will introduce a group of nature-inspired population based meta-heuristic optimization algorithms, including GA (Genetic Algorithm), SCE (Shuffled Complex Evolution), DE (Dierential Evolution), PSO (Particle Swarm Optimization), ES (Evolution Strategy) and CMA-ES (Covariance Matrix Adaption-Evolution Strategy). Different from the algorithms introduced in \cite{zhang2019derivativefree_part1}, which starts with one single solution candidate, the algorithms introduced in this part will start with a group of solution candidates candidates instead and propose to update them to improve the learning performance. When the objective variable space is too large to search exhaustively, the population based searches may be a good alternative, which cannot guarantee the optimal solution even though.

%----------------------------------------------------------------------------------------------

\subsection{Genetic Algorithm (GA)}\label{subsec:ga}

Genetic algorithm (GA) \cite{Whitley1994} is a meta-heuristic algorithm inspired by the process of natural selection in evolutionary algorithms, which has also been widely used for learning the solutions of many optimization problems. In GA, a population of candidate solutions will be initialized and evolved towards better ones. GA has demonstrated its outstanding performance in many learning scenarios, like non-convex objective function containing multiple local optima, objective function with non-smooth shape, as well as a large number of parameters and noisy environments. GA consists of several main steps, including \textit{generation initialization}, \textit{crossover and mutation}, \textit{fitness evaluation and selection}, which can effectively evolve good candidate solutions generation by generation. In this part, we will introduce these three main steps in great detail.

\subsubsection{Population Initialization}

Given the variable search space $\Theta$, a group of candidate solutions to the objective function can be generated via either random sampling from the space or the output from other existing learning algorithms if {\gadam} \cite{gadam} is used as the optimization framework (as introduced in the previous tutorial article \cite{zhang2019gradient}). Formally, we can denote the initial set of candidate solutions sampled from $\Theta$ as $\mc{G}^{(0)} = \{\boldsymbol{\theta}_1^{(0)}, \boldsymbol{\theta}_2^{(0)}, \cdots, \boldsymbol{\theta}_p^{(0)}\}$, where $p$ denotes the population size and the superscript denotes the generation index. These solution vectors are treated as the chromosome, which can be evolved to achieve better solutions. Traditional GA works well for the binary variable case, and several works also propose to extend GA to the real-number variable scenarios. Depending on the objective variable search space, the variable vectors $\boldsymbol{\theta}_i^{(0)} \in \Theta$ can contain either binary or real codes, and their corresponding sampling approaches can be different as well. 

For the binary variable search space, i.e., $\Theta \subseteq \{0, 1\}^{d_{\theta}}$, the entries in vector $\boldsymbol{\theta}_i^{(0)}$ can be sampled via the Bernoulli distribution, i.e., $\boldsymbol{\theta}_i^{(0)}(j) \sim \mc{B}(p), \forall j \in \{1, 2, \cdots, d_{\theta}\}$, where $p$ denotes the probability to sample value $1$. Meanwhile, for the real number variable search space, i.e., $\Theta \subseteq \mathbbm{R}^{d_{\theta}}$, the entries in vector $\boldsymbol{\theta}_i^{(0)}$ can be sampled via distributions like the Gaussian distribution, i.e., $\boldsymbol{\theta}_i^{(0)}(j) \sim \mc{N}(\mu, \sigma^2), \forall j \in \{1, 2, \cdots, d_{\theta}\}$, where $\mu$ and $\sigma$ denote the mean and standard deviation parameters of the distribution.

\subsubsection{Fitness Evaluation and Selection}

Among all the candidate solutions in set $\mc{G}$, some of them are good candidates for the objective problem but some of them can be not. GA will evaluate the candidate solutions in  $\mc{G}$ to pick the good ones as the parents to generate the offsprings. For the objective function mentioned in the Introduction section, we can evaluate the candidate solutions with the objective function $\mc{L}(\cdot)$ to be minimized. Formally, by applying the objective function on candidate solution $\boldsymbol{\theta}^{(0)}_i$, we can denote the introduced function value as $\ell_i^{(0)} = \mc{L}(\boldsymbol{\theta}^{(0)}_i)$. The loss terms introduced by function $\mc{L}(\cdot)$ on all the candidate solutions can be denoted as a list $[\ell_1^{(0)}, \ell_2^{(0)}, \cdots, \ell_p^{(0)}]$.

Generally, the solutions leading to smaller function values will have a larger chance to be selected in GA. There exist different ways to define the selection probability of each solution candidate. For instance, we can adopt the softmax equation to define the selection probability for $\boldsymbol{\theta}^{(0)}_i$ as follows:
\begin{equation}
p_i^{(0)} = \frac{ \exp(-\ell_i^{(0)}) }{\sum_{j = 1}^{p} \exp(-\ell_j^{(0)}) }.
\end{equation}
The selection probability of all the candidate solutions can be denoted as a list $[p_1^{(0)}, p_2^{(0)}, \cdots, p_p^{(0)}]$, based on which, $\frac{p}{2}$ pairs of unit model pairs will be selected as the parents for the next generation, which can be denoted as a set $\mc{P}^{(0)} = \left\{(\boldsymbol{\theta}_{i_1}^{(0)}, \boldsymbol{\theta}_{j_1}^{(0)}), (\boldsymbol{\theta}_{i_2}^{(0)}, \boldsymbol{\theta}_{j_2}^{(0)}), \cdots \right\}$, where $i_1, j_1, \cdots  \in \{1, 2, \cdots, p\}$.

%------------------------
\begin{figure}[t]
    \centering
    \includegraphics[width=0.9\textwidth]{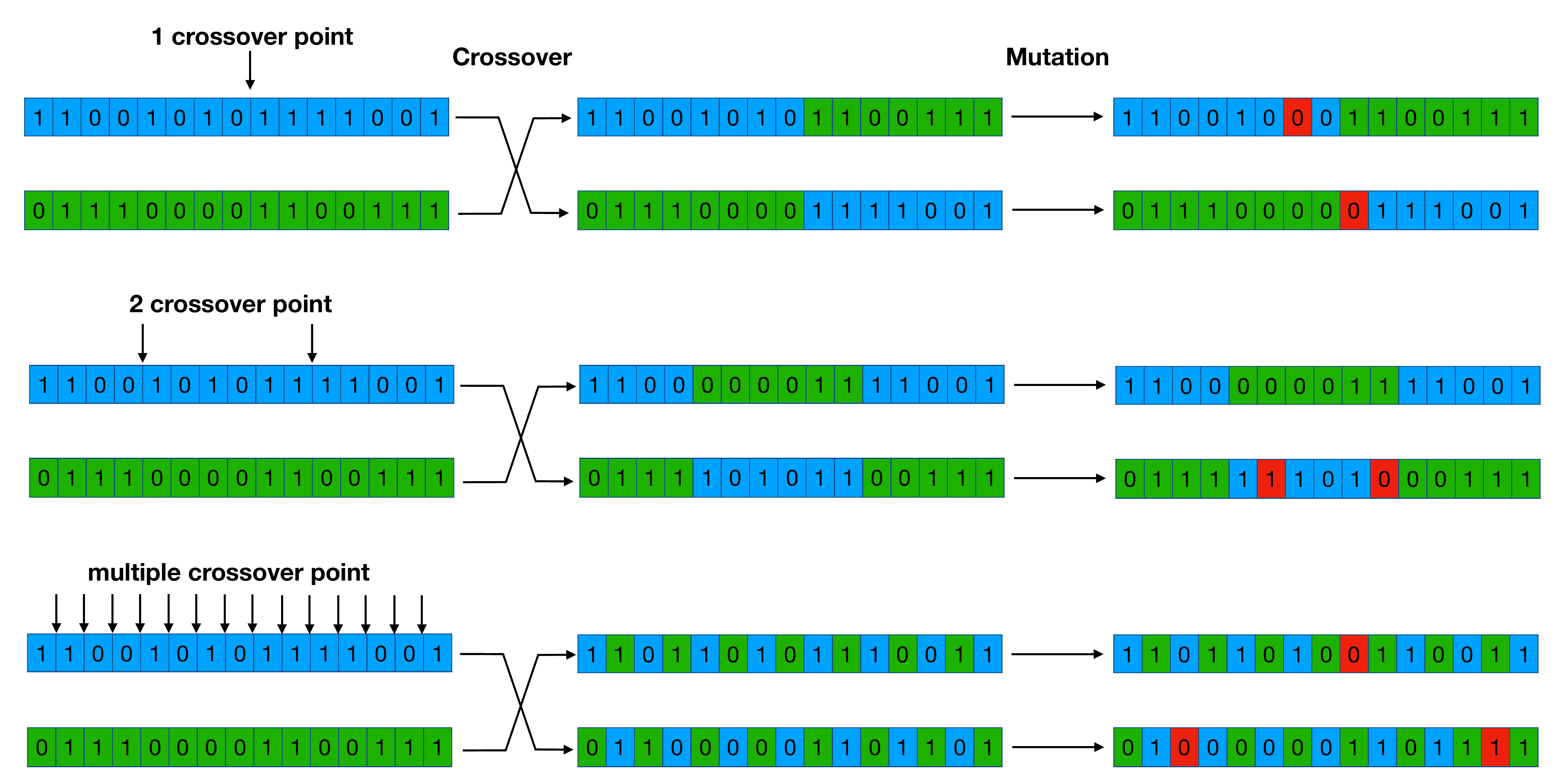}
    \caption{A Example of Crossover and Mutation Operations in GA.}
    \label{fig:chap_derivative_free_ga_crossover}
\end{figure}
%------------------------

\subsubsection{Crossover and Mutation}

GA generates the offsprings of the selected parent pairs via the crossover and mutation operations, which imitate the chromosome crossover and mutation of creatures in the natural world. Different kinds of crossover and mutation methods have been proposed already, and in this part we will introduce the classic crossover and mutation operations used in GA respectively.

\noindent \textbf{Crossover}: Given a parent variable pair, e.g., $(\boldsymbol{\theta}_{i_k}^{(0)}, \boldsymbol{\theta}_{j_k}^{(0)}) \in \mc{P}$, by selecting one or several crossover points, crossover aims at mix the variable values together to generate new children variables. For instance, as shown in Figure~\ref{fig:chap_derivative_free_ga_crossover}, we can represent the parent variable pair as the two sequences on the left-hand side (in blue and green colors respectively). By selecting the crossover point(s), GA will exchange the sub-sequences of the variable values between the parents to generate the children variable sequences, i.e., the ones on the right-hand side. The crossover points are usually selected by random actually. Depending on the number of crossover points finally selected, the crossover operation will create different offspring variables. In the plot, we show the examples with (1) one crossover point, (2) two crossover points, and (3) multiple crossover points, respectively, on the binary variables. Similar operations can be done on real-number variables as well.

\noindent \textbf{Mutation}: GA models the gene mutation in the real-world with the mutation operation, which can change a small number of the children variable values with a certain pre-specified probability. Formally, given the mutation probability $\eta \in [0, 1.0]$, GA will enumerate all the variable positions and randomly select a number of them subject to the probability $\eta$ for mutation. For instance, in Figure~\ref{fig:chap_derivative_free_ga_crossover}, for the generated children variables, a number of mutation spots are identified (in red color), the variable values at which are flipped (i.e., $1$ changed to $0$, and $0$ changed $1$). Some of the children variables have one single mutation spot, some have none and some may have two mutation spots. When it comes to the real-number variables, the variable mutation can be done in a similar way, where the bit flip operation can be replaced with some real-number sampling operation instead to change the variable values.

With the crossover and mutation operations, GA will generate a new group of candidate variable solutions, which can be denoted as set $\mc{G}^{(1)} = \{\boldsymbol{\theta}_1^{(1)}, \boldsymbol{\theta}_2^{(1)}, \cdots, \boldsymbol{\theta}_p^{(1)}\}$. Such an iterative process continues until convergence and the optimal variable in the last generation will be outputted as the solution. The pseudo-code of GA is provided in Algorithm~\ref{alg:ga}.

%------------------------------------------------------------------
%\setlength{\textfloatsep}{0pt}
\begin{algorithm}[t]
\small
\caption{Genetic Algorithm}
\label{alg:ga}
\begin{algorithmic}[1]
	\REQUIRE Variable Search Space $\Theta$.
\ENSURE  Model Parameter $\boldsymbol{\theta}$
\STATE	{Initialize a population $\mc{G} = \{\boldsymbol{\theta}_1, \boldsymbol{\theta}_2, \cdots, \boldsymbol{\theta}_p \}$}
\STATE	{Initialize convergence $tag = False$}
\WHILE	{$tag = False$}
\STATE	{Evaluate the solutions in $\mc{G}$ to get the loss function values $[\ell_1, \ell_2, \cdots, \ell_p]$}
\STATE	{Compute the sampling probability $[p_1^{(0)}, p_2^{(0)}, \cdots, p_p^{(0)}]$}
\STATE	{Sample the parent variable pairs $\mc{P} = \left\{(\boldsymbol{\theta}_{i_1}, \boldsymbol{\theta}_{j_1}), (\boldsymbol{\theta}_{i_2}, \boldsymbol{\theta}_{j_2}), \cdots \right\}$ subject to the probabilities}
\STATE	{Generate the children variables $\mc{G}'$ from pairs in $\mc{P}$ via crossover}
\STATE	{Update the children variables $\mc{G}''$ from $\mc{G}'$ via mutation}
\STATE	{Set $\mc{G} = \mc{G}''$}
\IF		{convergence condition holds}
\STATE	{$tag = True$}
\ENDIF
\ENDWHILE
\STATE	{Return $\boldsymbol{\theta}^* = \arg \min_{\boldsymbol{\theta} \in \mc{G}} \mc{L}(\bs{\theta})$}
\end{algorithmic}
\end{algorithm}
%------------------------------------------------------------------

\subsubsection{More Discussions on GA}
%------------------------
\begin{figure}[t]
    \centering
    \includegraphics[width=0.9\textwidth]{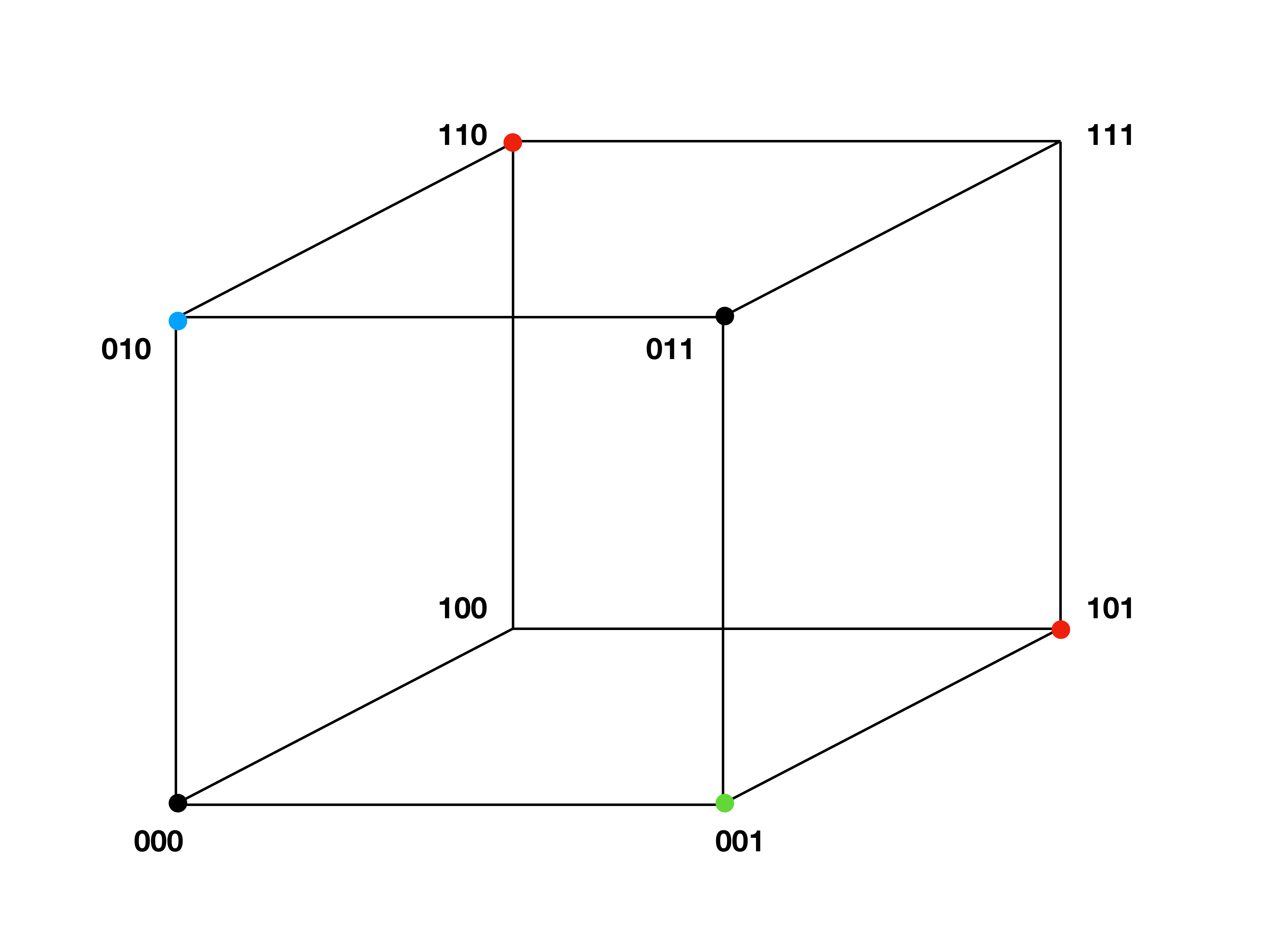}
    \caption{A Example to Illustrate the Physical Meanings of Crossover and Mutation in Optimization.}
    \label{fig:chap_derivative_free_ga_crossover_physical_meaning}
\end{figure}
%------------------------

Before we conclude this subsection on GA, we would like to provide the physical meanings of the crossover and mutation operations. Crossover actually will relocate the variable positions in the search space, which allows the algorithm to explore broadly based on the current feasible solutions. Meanwhile, to enable the algorithm to explore some regions outside the hyper-cubes with current solutions as the vertices, GA adopts the mutation operation which can change a small number of variable values in the learning process. 

For instance, in Figure~\ref{fig:chap_derivative_free_ga_crossover_physical_meaning}, we provide an example on optimization on a search space $\{0, 1\}^3$, where the variables are a binary sequence of length $3$ and the space include all the $8$ vertices of the cube. Let's assume, we are provided with two random variables $000$ and $011$, which correspond to the two vertices on the front-side of the cube. By using these two variables as the parents, via the crossover operation, these two vertices can generate the two children variables $010$ and $001$ (i.e., the remaining two vertices in blue and green colors respectively at the front-side). 

Here, we face a dilemma: no matter how we crossover the variables, we cannot explore the remaining $4$ vertices at the back-side. However, we discover that the mutation operation allows GA to resolve such a problem actually. For instance, by flipping the first bit of the variables from $0$ to $1$ in the children variables, GA will successfully reaches the back-side of the cube (i.e., the two red points) and will be able to explore the remaining vertices for learning the optimal solutions to the problem.

Traditional GA is usually very slow, and the learning process may involve a large number of generations. In recent years, some works have been introduced to improve the slow convergence problem of GA, including {\gadam} \cite{gadam} and fast GA \cite{DBLP:journals/corr/DoerrLMN17}. {\gadam} \cite{gadam} adopts the gradient descent algorithms to help reach the local optimum for each unit variable solutions before the evolution; while fast GA \cite{DBLP:journals/corr/DoerrLMN17} adopts a random mutation rate to enable the GA to achieve a faster convergence rate. {\gadam} is also introduced in our previous tutorial article \cite{zhang2019gradient}, and the readers may refer to the cited articles for more detailed information about these two mentioned algorithms.

%----------------------------------------------------------------------------------------------

\subsection{Shuffled Complex Evolution ({\sce}) Algorithm}

The Shuffled Complex Evolution ({\sce}) algorithm \cite{Duan1993} to be introduced in this part is an improvement over the genetic algorithm, which is based on a synthesis of four concepts that have been proven to be successful for global optimization, including (1) combination of probabilistic and deterministic approaches, (2) clustering, (3) systematic evolution of a complex of points spanning the search space, and (4) competitive evolution. {\sce} is shown to be much more effective, efficient and robust for a broad class of problems. In the following part of this subsection, we will introduce the outline of the {\sce} algorithm, together with the CCE (Competitive Complex Evolution) algorithm which will be used in {\sce}.

\subsubsection{{\sce} Algorithm Outline}

%------------------------------------------------------------------
%\setlength{\textfloatsep}{0pt}
\begin{algorithm}[t]
\small
\caption{{\sce} Algorithm}
\label{alg:sce_algorithm}
\begin{algorithmic}[1]
	\REQUIRE Variable search space $\Theta$; Complex number $p$; Complex size: $m$.
\ENSURE  Model Parameter $\boldsymbol{\theta}$
\STATE	{Compute sample size $s = p \times m$}
\STATE	{Sample $s$ variable points $\{\bs{\theta}_1, \bs{\theta}_2, \cdots, \bs{\theta}_s\}$ from space $\Theta$}
\STATE	{Evaluate the objective function $\mc{L}(\cdot)$ at the variables, and get $\{\ell_i = \mc{L}(\bs{\theta}_i)\}_{i \in \{1, 2, \cdots, s\}}$}
\STATE	{Sort the variables into an array $\mc{D} = \{\bs{\theta}_i\}_{i \in \{1, 2, \cdots, s\}}$ in an increasing order of $\{\ell_i\}_{i \in \{1, 2, \cdots, s\}}$}
\STATE	{Initialize convergence $tag = False$}
\WHILE	{$tag = False$}
\STATE	{Partition $\mc{D}$ sequentially into $p$ equal-sized complexes $\mc{A} = \{\mc{A}^1, \mc{A}^2, \cdots, \mc{A}^p\}$}
\FOR	{each complex $\mc{A}^i \in \mc{A}$}
\STATE	{Evolve $\mc{A}^i$ with the CCE algorithm, i.e., $\mc{A}^i = CCE(\mc{A}^i)$}
\STATE	{Add all the complexes in $\mc{A}$ into $\mc{D}$ again, i.e., $\mc{D} = \bigcup_{i = 1}^p \mc{A}^i$}
\STATE	{Evaluate variables in $\mc{D}$, and get $\{\ell_i = \mc{L}(\bs{\theta}_i)\}_{\bs{\theta}_i \in \mc{D}}$}
\STATE	{Resort variables in $\mc{D}$ in an increasing order of $\{\ell_i\}_{\bs{\theta}_i \in \mc{D}}$}
\ENDFOR
\IF		{convergence condition holds}
\STATE	{$tag = True$}
\ENDIF
\ENDWHILE
\STATE	{Return $\bs{\theta}^* = \arg \min_{\bs{\theta} \in \mc{D}} \mc{L}(\bs{\theta})$}
\end{algorithmic}
\end{algorithm}
%------------------------------------------------------------------

The outline of the {\sce} algorithm is illustrated in Algorithm~\ref{alg:sce_algorithm}, which accepts $p \ge 1$, $m \ge d_{\theta}+1$ and the search space $\Theta$ as the input. The algorithm consists of several important steps, which are listed as follows:
\begin{itemize}
\item \textbf{Initialization}: In the initialization step (i.e., line 1), {\sce} will compute the sample size $s = p \times m$.
\item \textbf{Sample Generation}: At line 2, $s$ variable samples will be sampled from the search space $\Theta$ (subject to the uniform distribution if no prior information is available).
\item \textbf{Evaluation and Sorting}: For each of the sampled variable point, {\sce} will evaluate the objective function at them. Considering our objective function is a minimization problem, the evaluation results will be used to sort the variable points in an increasing order of their function values (as indicated in lines 3-4).
\item \textbf{Complexes Partition}: The learning process of {\sce} involves an iterative process, which starts with a complex partition step as indicated in line 7. According to the increasing order of the points in $\mc{D}$, {\sce} sequentially partition these points into $p$ complexes $\mc{A} = \{\mc{A}^1, \mc{A}^2, \cdots, \mc{A}^p\}$, each of which contains $m$ points.
\item \textbf{Complex Evolution}: {\sce} will call the CCE algorithm to evolve the points in each complex to achieve better offspring variable solutions (i.e., in line 9). The CCE algorithm will be introduced later.
\item \textbf{Shuffle, Re-evaluate and Re-sort}: Based on the evolved variables, {\sce} will merge them together and re-evaluate the objective functions at these new points, which will be re-sorted again for the next iteration.
\end{itemize}
Such a process continues until convergence or the maximum iteration number has been met. The optimal variable in the last iteration will be outputted as the final solution, i.e., $\bs{\theta}^* = \arg \min_{\bs{\theta} \in \mc{D}} \mc{L}(\bs{\theta})$.

\subsubsection{CCE Algorithm Outline}

The CCE algorithm called in the {\sce} algorithm can effectively evolve the variable points in each complex into a new stage in a manner similar to the genetic algorithm, which accepts a complex $\mc{A}^k$, parameters $2 \le q \le m$, $\alpha \ge 1$ and $\beta \ge 1$ as the inputs. The output of CCE will be the evolved complex in a similar data structure as the input $\mc{A}^k$, containing the well-sorted variables. The pseudo-code of the CCE algorithm is provided in Algorithm~\ref{alg:cce_algorithm}.

The CCE algorithm consists of two evolution iterations. The external iteration of CCE samples the parent variable points, which will be evolved with the internal iteration. The key steps in CCE are introduced as follows:

%------------------------------------------------------------------
%\setlength{\textfloatsep}{0pt}
\begin{algorithm}[t]
\small
\caption{CCE Algorithm}
\label{alg:cce_algorithm}
\begin{algorithmic}[1]
	\REQUIRE Search space $\Theta$; Complex $\mc{A}^k$; Parent size $q$; Evolution iteration parameters $\alpha$ and $\beta$; .
\ENSURE  Evolved complex $\mc{A}^k$
\FOR	{external-evolution iteration index in $\{1, 2, \cdots, \beta\}$}
\STATE	{Compute weights for points in complex $\mc{A}^k$, i.e., $\{p_i\}_{\bs{\theta}^k_i \in \mc{A}^k}$}
\STATE	{Randomly select $q$ parent points $\mc{B} = \{\mb{u}_1, \mb{u}_2, \cdots, \mb{u}_q\}$ from $\mc{A}^k$ subject to the probabilities}
\FOR	{inner-evolution iteration index in $\{1, 2, \cdots, \alpha\}$}
\STATE	{Sort the points in $\mc{B}$ in the order of increasing function value}
\STATE	{Computing the centroid $\mb{g}$ of points in $\mc{B}$ excluding the worst point}
\STATE	{Reflect the worst point with a new point $\mb{r} = 2\mb{g} - \mb{u}_q$}
\IF		{$\mb{r}$ is not within $\Theta$}
\STATE	{Compute the smallest hypercube $\mc{H} \subset \mathbbm{R}^{d_{\theta}}$}
\STATE	{Randomly sample point $\mb{z}$ from $\mc{H}$}
\STATE	{Set $\mb{r} = \mb{z}$}
\ENDIF
\STATE	{Evaluate function $\mc{L}(\cdot)$ to get $\ell_r = \mc{L}(\mb{r})$ and $\ell_q = \mc{L}(\mb{u}_q)$}
\IF		{$\ell_r < \ell_q$}
\STATE	{Set $\mb{u}_q = \mb{r}$}
\ELSE	
\STATE	{Compute $\mb{c} = \frac{(\mb{g} + \mb{u}_q)}{2}$, and evaluate $\ell_c =\mc{L}(\mb{c})$}
\IF		{$\ell_c < \ell_q$}
\STATE	{Set $\mb{u}_q = \mb{c}$}
\ELSE
\STATE	{Randomly generate a point $\mb{z}$ from $\mc{H}$}
\STATE	{Set $\mb{u}_q = \mb{z}$}
\ENDIF
\ENDIF
\ENDFOR
\STATE	{Replace the original points in $\mc{A}^k$ with the evolved ones in $\mc{B}$}
\STATE	{Resort $\mc{A}^k$ according to the re-evaluated function values at points in $\mc{A}^k$}
\ENDFOR
\STATE	{Return $\mc{A}^k$ as the output}
\end{algorithmic}
\end{algorithm}
%------------------------------------------------------------------

\noindent \underline{\textbf{External Evolution Iteration for $\beta$ Rounds}}
\begin{itemize}
\item \textbf{Weight Computation}: Based on the position indexes of points in complex $\mc{A}^k$, CCE will compute the sampling probabilities for these points, which can be denoted as
\begin{equation}
p_i = \frac{2(m+1-i)}{m(m+1)}, \forall i \in \{1, 2, \cdots, m\},
\end{equation}
where $i$ here denotes the index of a variable point. In other words, for the points which are in the front (i.e., with smaller function evaluation values) will have a higher sampling opportunity. For instance, for the first point in $\mc{A}^k$, its sampling probability will be $\frac{2}{(m+1)}$; while for the last point in $\mc{A}^k$, its sampling probability will be $\frac{2}{m(m+1)}$.

\item \textbf{Parent Selection}: A batch of points will be randomly selected from $\mc{A}^k$ as the parents, which is denoted as $\mc{B} = \{\mb{u}_1, \mb{u}_2, \cdots, \mb{u}_q\}$ in line 3. Here, we use the notation $\mb{u}$ instead of $\bs{\theta}$ to avoid the confusion about the subscripts in the representation.

\item \textbf{\underline{Internal Evolution Iteration}: Offspring Generation for $\alpha$ Rounds}
\begin{itemize}
\item \textbf{Centroid Computing}: Variable points in the parent batch $\mc{B}$ will be sorted, which will all (except the worst point) be used to compute the centroid of the complex, i.e.,
\begin{equation}
\mb{g} = \frac{1}{q-1} \sum_{i = 1}^{q-1} \mb{u}_i.
\end{equation}
\item \textbf{Reflection Step}: A new point will be computed based the worst point in $\mc{B}$, which can be denoted as $\mb{r} = 2\mb{g} - \mb{u}_q$.
\item \textbf{Mutation Step}: If the new point $\mb{r}$ is not in the search space $\Theta$, CCE will replace $\mb{r}$  with a new randomly selected point $\mb{z}$ from the smallest hypercube $\mc{H}$ which covers all the points in $\mc{A}^k$.
\item \textbf{Contraction Step}: In the case if $\mb{r}$ is better than $\mb{u}_q$, CCE will replace $\mb{u}_q$ with $\mb{r}$; otherwise, CCE will create a new point $\mb{c} = \frac{(\mb{g} + \mb{u}_q)}{2}$ (i.e., the central point between centroid $\mb{g}$ and the worst point $\mb{u}_q$).
\item \textbf{Mutation Step}: If $\mb{c}$ is better than $\mb{u}_q$, CCE will replace $\mb{u}_q$ with $\mb{c}$; otherwise, CCE will replace $\mb{u}_q$ with a randomly selected point $\mb{z} \in \mc{H}$.
\end{itemize}

\item \textbf{Parent Update}: All the generated offsprings will be put back into $\mc{A}^k$ to update the variables. Complex $\mc{A}^k$ will be updated, re-evaluated, and re-sorted for the next iteration. 
\end{itemize}

\subsubsection{More Discussions on {\sce}}

The {\sce} algorithm treats the global search of optimal solutions as a process of natural evolution, where the sampled $s$ points contribute a population. The population will be partitioned into several communities, each of which will evolve independently (which denotes the process to search the space in different directions). After a certain number of generations, the communities will be forced to mix together, and new communities will be created via a shuffling process. This procedure enhances survivability by a sharing of information gained independently by each community.

The evolution process used in {\sce} is different from that in GA, where the parents are in a batch instead of a pair. A subset of the points will be sampled from a complex subject to the pre-computed probabilities, which serve as the parents in the evolution. The offsprings are introduced at random locations of the feasible search space under certain condition that the evolution will not be trapped by unpromising regions. Therefore, each mutation will help improve the community slightly in the evolution process, and the newly generated points will replace the worst point in the community. It is very important for the effectiveness of {\sce} in guiding the search process.

%----------------------------------------------------------------------------------------------

\subsection{Differential Evolution ({\de}) Algorithm}

Differential Evolution ({\de}) algorithm \cite{DE} is a new heuristic approach mainly having three advantages; finding the true global minimum regardless of the initial parameter values, fast convergence, and using few control parameters. DE algorithm is a population based algorithm like GA using similar operators; crossover, mutation and selection. Meanwhile, in searching for better solutions, traditional GA heavily rely on crossover operation for local search; while {\de} relies more on the mutation operation instead. {\de} utilizes mutation operation for the solution search purposes, and applies the selection operation to direct the search toward the prospective regions in the variable space. In this subsection, we will provide a brief introduction to the DE algorithm. The general algorithm framework of {\de} is very similar to GA, and we will not provide its pseudo-code here but focus on introducing the three main operations as follows.

\subsubsection{Mutation}

Formally, let $\mc{P}^{(k)} = \{\bs{\theta}_1^{(k)}, \bs{\theta}_2^{(k)}, \cdots, \bs{\theta}_p^{(k)}\}$ denote a set of variables evolved to the $k_{th}$ generation, where $p$ denotes the population size. The variables in the initial generation, i.e., $\mc{P}^{(0)}$, are sampled randomly from the search space $\Theta$ if no prior knowledge about the problem optimal solution is available. 

The crucial idea behind {\de} is the mutation scheme for generating trial variable vectors. {\de} generates new variable vectors by adding a weighted difference vector between two population members to a third member. If the resulting vector yields a lower objective function value than a predetermined population member, the newly generated vector will replace the vector with which it was compared in the following generation. The comparison vector can but need not be part of the generation process mentioned above. In addition, the best parameter vector $\bs{\theta}^{(k)}_{best}$ will be evaluated for every generation in order to keep track of the progress that is made during the minimization process. Several different mutation schemes are introduced in \cite{DE}, and we will introduce them as follows.

\noindent \textbf{Scheme 1:}
For each variable vector $\bs{\theta}^{(k)}_i \in \mc{P}^{(k)}$, {\de} will generate a trial vector $\mb{v}_i^{(k)}$ for it as follows:
\begin{equation}
\mb{v}_i^{(k)} = \bs{\theta}^{(k)}_{r_1} + F \cdot (\bs{\theta}^{(k)}_{r_2} - \bs{\theta}^{(k)}_{r_3}),
\end{equation}
where the indexes $r_1, r_2, r_3$ are randomly selected from $\{1, 2, \cdots, p\}$ and $r_1, r_2, r_3 \neq i$. Term $F$ is a real constant which controls the amplification of the differential variation term. 

\noindent \textbf{Scheme 2:}
Another scheme introduced in \cite{DE} is very similar to the above scheme 1, which further considers the best variable in the current generation when generating the trial vector $\mb{v}_i^{(k)}$ for the variables. Formally, let $\bs{\theta}^{(k)}_{best}$ denote the optimal variable in the current generation $\mc{P}^{(k)}$, which introduces the lowest objective function value. We can represent the generated vector for variable $\bs{\theta}_i^{(k)} \in \mc{P}^{(k)}$ as
\begin{equation}
\mb{v}_i^{(k)} = \bs{\theta}^{(k)}_{i} + \lambda \cdot (\bs{\theta}^{(k)}_{best} - \bs{\theta}^{(k)}_{i}) + F \cdot (\bs{\theta}^{(k)}_{r_2} - \bs{\theta}^{(k)}_{r_3}),
\end{equation}
where $\lambda$ is an additional introduced control parameter to control the greediness of the scheme by incorporating the current best vector $\bs{\theta}^{(k)}_{best}$. This new term can be extremely useful for objective functions where the global minimum is relatively easy to find.

\subsubsection{Crossover}

Based on the generated vector $\{\mb{v}_i^{(k)}\}_{i \in \{1, 2, \cdots, p\}}$ (with either scheme 1 or scheme 2), {\de} proposes to crossover it with the original variable vector $\bs{\theta}^{(k)}_{i}$. By randomly selecting the crossover index point $n$ as well as the crossover sequence length $L$, one segment of the variable values from vector $\mb{v}_i^{(k)}$ will be used to generate a new vector $\mb{u}_i^{(k)}$ of length $d_{\theta}$. To be more specific, the entries in vector $\mb{u}_i^{(k)}$ can be represented with the following equation:
\begin{equation}
\mb{u}_i^{(k)}(j) = \begin{cases}
\mb{v}_i^{(k)}(j), & \mbox{ if } j \in \{n, n+1, \cdots, n+L-1\},\\
\bs{\theta}_i^{(k)}(k), & \mbox{ otherwise}.
\end{cases}
\end{equation}

\begin{example}
%------------------------
\begin{figure}[t]
    \centering
    \includegraphics[width=0.5\textwidth]{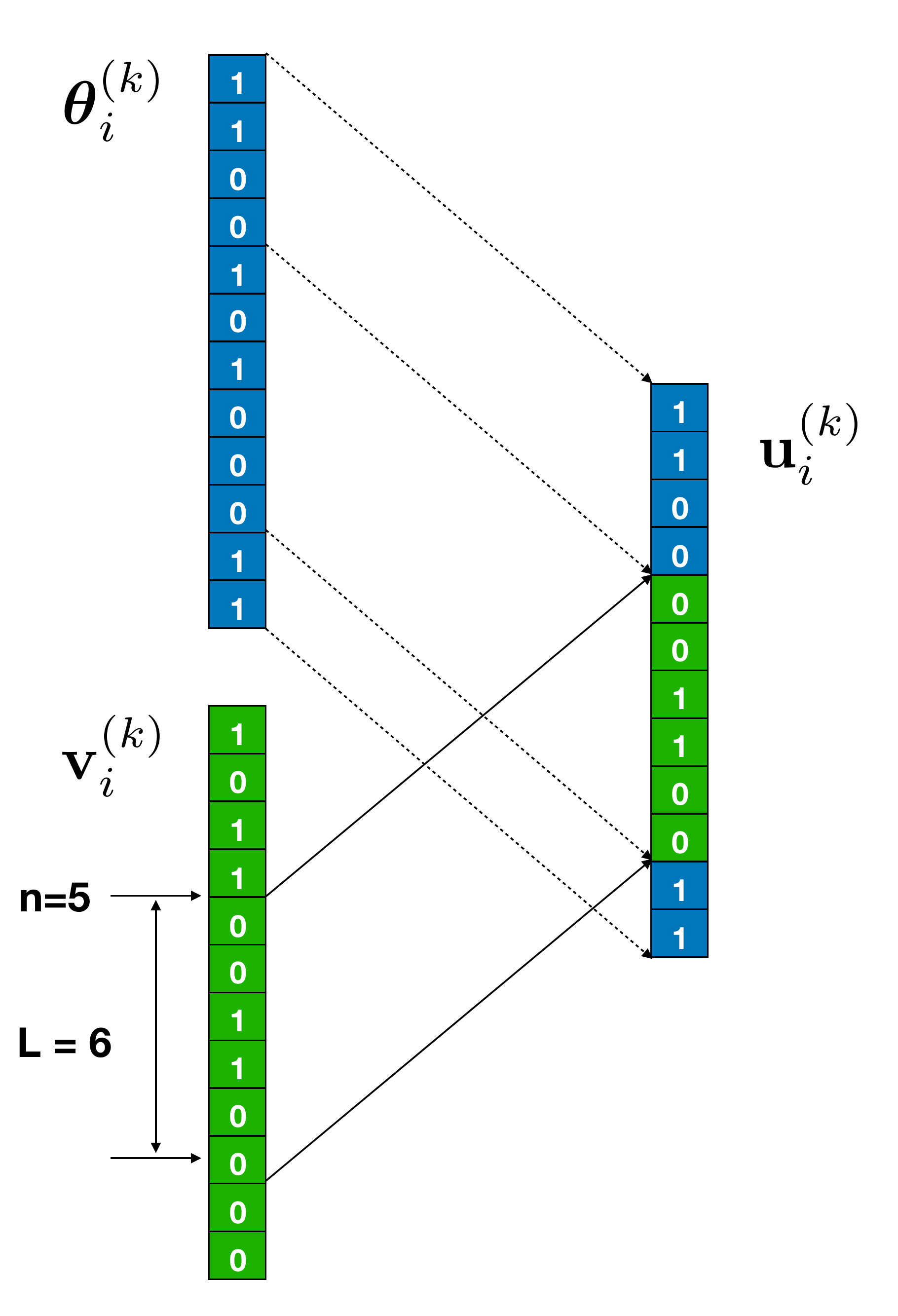}
    \caption{A Example of Crossover Operations in {\de}.}
    \label{fig:chap_derivative_free_de_crossover}
\end{figure}
%------------------------

For instance, as shown in Figure~\ref{fig:chap_derivative_free_de_crossover}, given the two variable vectors $\mb{v}_i^{(k)}$ (in green color) and $\bs{\theta}_i^{(k)}$ (in blue color), by sampling the crossover starting index $n$ and the crossover segment length $L$, {\de} generates a new vector $\mb{u}_i^{(k)}$ as shown at the right hand side. The entries $\mb{u}_i^{(k)}(5:10)$ are from $\mb{v}_i^{(k)}$, and the remaining entries in $\mb{u}_i^{(k)}$ are from $\bs{\theta}_i^{(k)}$ instead.
\end{example}

%------------------------------------------------------------------
%\setlength{\textfloatsep}{0pt}
\begin{algorithm}[H]
\caption{Sampling Approach of $L$}
\label{alg:L_sampling_code}
\small
\begin{algorithmic}[1]
\REQUIRE Crossover probability $p$; Variable dimension $d_{\theta}$; Starting index $n$
\ENSURE  Crossover segment length $L$
\STATE	{$L = 1$}
\WHILE	{$rand() < p \land (n+L) < d_{\theta}$}
\STATE	{$L += 1$}
\ENDWHILE
\STATE	{Return $L$}
\end{algorithmic}
\end{algorithm}
%------------------------------------------------------------------

In the {\de} algorithm, the starting index $n$ is usually sampled from set $\{1, 2, \cdots, d_{\theta}\}$ subject to the uniform distribution. {\de} determines the crossover segment length with the following pseudo-code in Algorithm~\ref{alg:L_sampling_code}.

In the algorithm, $p$ is the crossover probability and $rand()$ denotes a random number sampled in range $[0, 1]$ subject to the uniform distribution. According to the algorithm, the probability to get a crossover segment of length no less than $l$ will be $P(L \ge l) = p^{l-1}$.

\subsubsection{Selection}

The selection operation in {\de} is very simple. Based on the original variable $\bs{\theta}_i^{(k)}$ and the newly generated vector $\mb{u}_i^{(k)}$ for it, {\de} will select which one should be used in the next generation, i.e., generation $k+1$, with the following equation:
\begin{equation}
\bs{\theta}_i^{(k+1)} = \begin{cases}
\mb{u}_i^{(k)}, \mbox{ if } \mc{L}(\mb{u}_i^{(k)}) < \mc{L}(\bs{\theta}_i^{(k)}),\\
\bs{\theta}_i^{(k)}, \mbox{ otherwise}.
\end{cases}
\end{equation}
In other words, if vector $\mb{u}_i^{(k)}$ can lead to a smaller function value, it will be used to replace $\bs{\theta}_i^{(k)}$ in the next generation; otherwise, $\bs{\theta}_i^{(k)}$ value is retained in the next generation.

\subsubsection{More Discussions on {\de}}

Traditional direction search algorithms, including GA, actually adopt a greedy strategy for selecting good solutions, where variations are created on the solution vectors. Once a variation is generated, the algorithm will evaluate its benefits, which will be accepted as new solutions if it reduces the objective function value (for minimization problems). Such a kind of algorithm can converge fast but may get trapped into the local minima. {\de} resolves such a disadvantage by adopting a new mutation schema, which makes {\de} to be self-adaptive. In {\de}, all the solutions have the same chance to be selected as the parents without dependence of their fitness value. {\de} also adopts a greedy selection process: the better one of the new solution and its parent wins the competition providing significant advantages of converging performance over genetic algorithm. {\de} is a stochastic optimization algorithm, which can optimize the objective function and considering the constraints on the variables. Compared with other algorithms, {\de} has three advantages: (1) identifying true global minimum, (2) fast convergence, and (3) a few control parameters.

%----------------------------------------------------------------------------------------------

\subsection{Particle Swarm Optimization (PSO) Algorithm}

The {\pso} (Particle Swarm Optimization) algorithm mimics the social behavior of birds flocking and fishes schooling starting from a randomly distributed set of particles (i.e., the potential solution candidates). Similar to GA, the {\pso} algorithm will improve the solutions according to a quality measure (i.e., the fitness function) by moving the particles around the search space with a set of simple mathematical expressions. The {\pso} algorithm shares a lot of common elements with GA:
\begin{itemize}
\item Both initialize a population in a similar manner;
\item Both use an evaluation function to determine how fit a potential solution is;
\item Both are generational by repeating the same set of processes.
\end{itemize}

The {\pso} algorithm has two main operators: \textit{velocity update} and \textit{position update}. During each generation, each particle is accelerated toward the particle's previous best position and the global best position. In each generation, a new velocity value for each particle is calculated based on (1) its current velocity, (2) the distance from its previous best position, and (3) the distance from the global best position, which will be used to calculate the next position of the particle in the search space. The pseudo-code of the {\pso} algorithm is provided in Algorithm~\ref{alg:pso_algorithm}.

%------------------------------------------------------------------
%\setlength{\textfloatsep}{0pt}
\begin{algorithm}[t]
\small
\caption{{\pso} Algorithm}
\label{alg:pso_algorithm}
\begin{algorithmic}[1]
	\REQUIRE Variable search space $\Theta$; Particle population $g$.
\ENSURE  Model Parameter $\boldsymbol{\theta}$
\STATE	{Initialize a set of particles $\mc{P} = \{P_1, P_2, \cdots, P_g\}$}
\FOR	{each particle $P_i \in \mc{P}$}
\STATE	{Initialize $\boldsymbol{\theta}_i$, $\mb{v}_i$ and $\boldsymbol{\theta}^*_i$ from search space $\Theta$ for each particle $P_i$}
\ENDFOR
\STATE	{Initialize convergence $tag=False$}
\WHILE	{$tag == False$}
\FOR	{each particle $P_i \in \mc{P}$}
\STATE	{/*Evaluate objective function $\mc{L}(\boldsymbol{\theta}_i)$ and update the optimal variable $\boldsymbol{\theta}_i^*$*/}
\IF		{$\mc{L}(\boldsymbol{\theta}_i) < \mc{L}(\boldsymbol{\theta}_i^*)$}
\STATE	{Set the best solution found so far $\boldsymbol{\theta}_i^* = \boldsymbol{\theta}_i$}
\ENDIF
\STATE	{/*Set $g$ to be the index of the optimal neighbor of $P_i$*/}
\STATE	{Set $g = i$}
\FOR	{particle $P_j$ in $P_i$'s neighborhood set}
\IF		{$\mc{L}(\boldsymbol{\theta}_j^*) < \mc{L}(\boldsymbol{\theta}_g^*)$}
\STATE	{Update $g = j$}
\ENDIF
\ENDFOR
\STATE	{/*Update terms $\mb{v}_i$ and $\boldsymbol{\theta}_i$*/}
\STATE	{Update the velocity $\mb{v}_i$ term with the identified $g$, i.e., $\mb{v}_i = f(\boldsymbol{\theta}_i, \mb{v}_i, \boldsymbol{\theta}_i^*, \boldsymbol{\theta}_g^*)$}
\STATE	{Update the position $\boldsymbol{\theta}_i$ term, i.e., $\boldsymbol{\theta}_i = h(\boldsymbol{\theta}_i, \mb{v}_i)$}
\ENDFOR
\IF		{Convergence condition holds}
\STATE	{Set $tag = True$}
\ENDIF
\ENDWHILE
\STATE	{Return $\boldsymbol{\theta}^* = \arg \max_{i \in \{1, 2, \cdots, n\}} \mc{L}(\boldsymbol{\theta}_i)$}
\end{algorithmic}
\end{algorithm}
%------------------------------------------------------------------

According to the algorithm, in lines 1-5, the algorithm initializes a set of variables, including the particle set, their variable, velocity and optimal variable vectors, respectively. In the learning process, for each particle, the algorithm will evaluate the function at the current variable to update the best seen variable (i.e., lines 9-11) as well as identify the local optimal neighbor index (i.e., lines 13-18), which will be used to update both the velocity vector as well as the variable vector with equations shown in lines 20-21. In the following part, we will introduce the specific representations of the function $\mb{v}_i = f(\boldsymbol{\theta}_i, \mb{v}_i, \boldsymbol{\theta}_i^*, \boldsymbol{\theta}_g^*)$ and function $\boldsymbol{\theta}_i = h(\boldsymbol{\theta}_i, \mb{v}_i)$ proposed in the existing works for different scenarios.

\subsubsection{Binary PSO Algorithm}

In the case when the search space is binary, i.e., $\Theta = \{0, 1\}^{d_{\theta}}$, the corresponding {\pso} used for learning the variable will be called the binary {\pso} algorithm \cite{637339}. In the algorithm, the velocity vector $\mb{v}_i$ for particle $P_i$ keep records of the current velocity, whose value is determined by both the velocity in the previous round, the velocity of the historical best seen variable, as well as the optimal neighbor variable value. To differentiate the velocity vector in different iterations, we use $\mb{v}_i^{(\tau)}$ to denote the velocity in iteration $\tau$. For vector $\mb{v}_i^{(\tau)} = [{v}_i^{(\tau)}(1), {v}_i^{(\tau)}(2), \cdots, {v}_i^{(\tau)}(d_{\theta})]$, its $j_{th}$ entry $\mb{v}_i^{(\tau)}(j)$ will be updated with the following equation:
\begin{align}
\mb{v}_i^{(\tau)}(j) &= f(\boldsymbol{\theta}^{(\tau-1)}_i, \mb{v}^{(\tau-1)}_i, \boldsymbol{\theta}_i^*, \boldsymbol{\theta}_g^*)\\
&= \mb{v}_i^{(\tau-1)}(j) + c_1 \cdot \psi_1 \cdot \left(\boldsymbol{\theta}_i^*(j) - \boldsymbol{\theta}^{(\tau - 1)}_i(j) \right) + c_2 \cdot \psi_2 \cdot \left(\boldsymbol{\theta}_g^*(j) - \boldsymbol{\theta}^{(\tau - 1)}_i(j) \right),
\end{align}
where $c_1, c_2$ denote the two weight parameters (usually set with value $2.0$), and terms $\psi_1$, $\psi_2$ represent two random number drawn from a uniform distribution between $0.0$ and $1.0$. In many of the cases, there will exist a bound pair $[v_{min}, v_{max}]$ to constrain the possible values of vector $\mb{v}_i$ (values exceeding the bound will be smoothen accordingly).

Based on the updated velocity vector $\mb{v}_i^{(\tau)}$, the {\pso} algorithm will update the variable value vector $\boldsymbol{\theta}_i$. Generally, for the larger values in vector $\mb{v}_i^{(\tau)}$, the corresponding entry in vector $\boldsymbol{\theta}_i$ will be more likely to have value $1$. The binary {\pso} algorithm will update the variable entry $\boldsymbol{\theta}^{(\tau)}_i(j)$ with the following equation:
\begin{equation}
\boldsymbol{\theta}^{(\tau)}_i(j) = h(\boldsymbol{\theta}^{(\tau-1)}_i, \mb{v}^{(\tau)}_i) = \begin{cases}
1,& \mbox{ if } \psi_3 < \frac{1}{1+\exp(- \mb{v}_i^{(\tau)}(j))},\\
0,& \mbox{ otherwise;}
\end{cases}
\end{equation} 
where term $\psi_3$ is a random number selected with the uniform distribution from range $[0, 1.0]$.

\subsubsection{Standard PSO Algorithm}

In the case when the objective variables are real numbers, i.e., the search space is $\Theta = \mathbb{R}^{d_{\theta}}$, the variables will denote a point in the search space, and the {\pso} algorithm will be called the standard {\pso} algorithm \cite{swarm_intelligence}. Formally, in the standard {\pso} algorithm, the velocity and variable vectors will be updated with the following equations respectively:
\begin{align}
\mb{v}_i^{(\tau)}(j) &= \mb{v}_i^{(\tau-1)}(j) + c_1 \cdot \psi_1 \cdot \left(\boldsymbol{\theta}_i^*(j) - \boldsymbol{\theta}^{(\tau - 1)}_i(j) \right) + c_2 \cdot \psi_2 \cdot \left(\boldsymbol{\theta}_g^*(j) - \boldsymbol{\theta}^{(\tau - 1)}_i(j) \right);\\
\boldsymbol{\theta}_i^{(\tau)}(j) &= \boldsymbol{\theta}_i^{(\tau-1)}(j) + \mb{v}_i^{(\tau)}(j).
\end{align}

Similarly to the binary {\pso}, to avoid the oscillations of the velocity vector $\mb{v}_i^{(\tau)}$, there usually exist a tight lower and upper bounds $[v_{min}, v_{max}]$ for the entries, where the ones exceeding the boundaries will be smoothen effectively with values $v_{min}$ and $v_{max}$ respectively.

\subsubsection{PSO with Inertia}

In this part, we will introduce the {\pso} algorithm with inertia \cite{10.1007/BFb0040810}, which assigns the historical velocity with a weight in the velocity vector updating equation. For a nonzero weight, the algorithm will move the particles in the same direction as the previous iterations. Meanwhile, a decreasing weight over iterations will introduce a shift from the global search to the local search instead. Formally, the velocity and variable updating equation adopted in the {\pso} algorithm with inertia can be denoted as follows:
\begin{align}
\mb{v}_i^{(\tau)}(j) &= w^{(\tau)} \cdot \mb{v}_i^{(\tau-1)}(j) + c_1 \cdot \psi_1 \cdot \left(\boldsymbol{\theta}_i^*(j) - \boldsymbol{\theta}^{(\tau - 1)}_i(j) \right) + c_2 \cdot \psi_2 \cdot \left(\boldsymbol{\theta}_g^*(j) - \boldsymbol{\theta}^{(\tau - 1)}_i(j) \right);\\
\boldsymbol{\theta}_i^{(\tau)}(j) &= \boldsymbol{\theta}_i^{(\tau-1)}(j) + \mb{v}_i^{(\tau)}(j),
\end{align}
where term $w^{(\tau)}$ denotes the inertia weight. 

Generally, the weight term will be reduced linearly with iteration, from $w_{start}$ to $w_{end}$ ($w_{start}$ and $w_{end}$ are usually set with values $0.9$ and $0.4$ respectively). The updating equation of term $w^{(\tau)}$ can be represented as follows
\begin{equation}
w^{(\tau)} = \frac{(T_{max} - \tau) \cdot (w_{start} - w_{end})}{T_{max}} + w_{end},
\end{equation}
where $T_{max}$ denotes the maximum iteration round allowed in the {\pso} algorithm. A the algorithm continues, the value of $w^{(\tau)}$ will gradually reduces from $w_{start}$ to $w_{end}$.

\subsubsection{PSO with Constriction Coefficient}

Another {\pso} variant \cite{785513} to be introduced in this section incorporates the constriction coefficient into the velocity updating equation, which will lead to particle convergence over iterations, since the amplitude of the particle's oscillations decrease as it focus on the local and neighborhood previous best solutions. Meanwhile, the constriction coefficient also prevents colapse if the right social conditions are in place. The particle will oscillate around the weighted mean of the historical best solution and the optimal neighbor's best solution if they are near each other, which performs a local search in the space. On the other hand, if these solutions are far away from each other, the {\pso} algorithm will perform a global search instead. The constriction coefficient will balance between local search and global search effectively depending on the constriction coefficient value.

The updating equations of the velocity and variable vectors in the {\pso} with constriction coefficient algorithm can be represented as follows:
\begin{align}
\mb{v}_i^{(\tau)}(j) &= \chi \cdot \left[ \mb{v}_i^{(\tau-1)}(j) + c_1 \cdot \psi_1 \cdot \left(\boldsymbol{\theta}_i^*(j) - \boldsymbol{\theta}^{(\tau - 1)}_i(j) \right) + c_2 \cdot \psi_2 \cdot \left(\boldsymbol{\theta}_g^*(j) - \boldsymbol{\theta}^{(\tau - 1)}_i(j) \right) \right];\\
\boldsymbol{\theta}_i^{(\tau)}(j) &= \boldsymbol{\theta}_i^{(\tau-1)}(j) + \mb{v}_i^{(\tau)}(j),
\end{align}
where $\chi = \frac{2k}{|2 - \psi - \sqrt{\psi^2 - 4\psi} |}$ and $\psi = c_1 + c_2$. Normally, variable $k$ is set with value $1$ and $c_1, c_2$ are assigned with value $2$ in the algorithm. 

%----------------------------------------------------------------------------------------------

\subsection{Evolution Strategy (ES) Algorithm}

Evolution Strategy ({\es}) \cite{es}, also known as Evolutionary Strategy, is a search paradigm inspired by the principles of the biological evolution, which also belongs to the population based evolutionary algorithms. {\es} and GA work in a very similar way, involving selection, mutation and crossover (called recombination in {\es}), which were developed independent by two groups of  researchers ({\es} was developed by the European computer scientists and GA was introduced the the USA computer scientists). In most of the cases, GA adopts a binary code for the solutions, while {\es} works well for the optimization functions with real-number variables. A more detailed discussions on the differences and similarities between {\es} and GA is available in \cite{10.1007/BFb0029787}. In this subsection, we will first talk about the general algorithm framework of {\es}, where the detailed parameter control strategies used in {\es} will be introduced in the following subsection.

\subsubsection{Algorithm Outline}

The {\es} algorithm involves an iterative procedure, where new individuals will be created in each generation from the existing ones. Formally, to specify the learning settings of the algorithm, {\es} can be normally denoted as $(\mu/\rho \begin{smallmatrix}+\\,\end{smallmatrix} \lambda)$-{\es}. In the case where parameter $\rho$ is not specified, the {\es} algorithm can also be denoted as $(\mu \begin{smallmatrix}+\\,\end{smallmatrix} \lambda)$-{\es} as well. Here, $\mu$, $\rho$ and $\lambda$ are positive integers, whose physical meanings are denoted as follows:
\begin{itemize}
\item $\mu$: the number of individuals in the parent set, i.e., the parent population.
\item $\rho$: the number of parent individuals selected (out of $\mu$ parents) for recombination.
\item $\lambda$: the number of offsprings generated in each iteration.
\item $\begin{smallmatrix}+\\,\end{smallmatrix}$: the two selection modes of the algorithm. If notation ``$+$'' is used (i.e., the ``plus'' selection mode), the individual age is not considered in selection, and $\mu$ best of the $\mu+\lambda$ individuals will be selected as parents in each iteration. If notation ``$,$'' is used (i.e., the ``comma'' selection mode), the senior individuals will die out after each iteration, and $\mu$ out of the generated $\lambda$ offsprings will be selected as parents in each iteration.
\end{itemize}
In the $(\mu , \lambda)$-{\es}, $\mu \le \lambda$ should always hold; while in the $(\mu + \lambda)$-{\es}, $\mu \le \lambda$ is not necessary any more and $\lambda = 1$ is also a feasible setting for the algorithm. In some cases, a subscript will be attached to $\rho$ to denote the combination modes: $\rho_\textsc{i}$ for intermediate recombination and $\rho_\textsc{w}$ for weighted recombination. Intermediate recombination is also the default recombination mode if the subscript is not indicated. 

The pseudo-code of the $(\mu/\rho \begin{smallmatrix}+\\,\end{smallmatrix} \lambda)$-{\es} algorithm is provided in Algorithm~\ref{alg:es_algorithm}, which accepts $d_{\theta}$, $\rho$, $\mu$ and $\lambda$ as the input. At the beginning, the algorithm will initialize an individual $\bs{\theta}$ at the beginning. The algorithm will also create a parameter variable $s$ for it, which can covers control or endogenous strategy parameters, e.g., the success counter or a step-size that primarily serves to control the mutation. As the algorithm continues, a group of $\lambda$ offspring instances will be generated via the mutation operations, where the $\mu$ good instances will be selected to form the new parent set. The initial variable $\bs{\theta}$ and parameter $s$ will be updated with the selection and recombination operations. Detailed information about the operations used in the pseudo-code will be introduced in detail as follows.

%------------------------------------------------------------------
%\setlength{\textfloatsep}{0pt}
\begin{algorithm}[t]
\small
\caption{Evolution Strategy Algorithm}
\label{alg:es_algorithm}
\begin{algorithmic}[1]
	\REQUIRE Variable search space $\Theta$; Variable dimension $d_{\theta}$; Evolution parameters $\rho$, $\mu$ and $\lambda$
\ENSURE  Model Parameter $\boldsymbol{\theta}$
\STATE	{Initialize parent population $\mc{P} = \{\}$}
\STATE	{Initialize an instance with variable vector $\bs{\theta} \in \mathbbm{R}^{d_{\theta}}$ with control parameter $s$}
\STATE	{Initialize convergence $tag = False$}
\WHILE	{$tag = False$}
\FOR	{$i \in \{1, 2, \cdots, \lambda\}$}
\STATE	{$\bs{\theta}_i$, $s_i$ = mutate($\bs{\theta}$, $s$)}
\STATE	{$\mc{P} = \mc{P} \cup \left\{( \bs{\theta}_i, s_i, \mc{L}(\bs{\theta}_i) )\right\}$}
\ENDFOR
\STATE	{$\mc{P}$ = select\_by\_age($\mc{P}$)}
\STATE	{$\mc{P}$ = select\_$\mu$\_best($\mu$, $\mc{P}$) \ \ \ \ // ** optional ** }
\STATE	{($\bs{\theta}$, $s$) = recombine(select\_mate($\mc{P}$, $\rho$), $\bs{\theta}$, $s$)}
\IF		{Convergence condition holds}
\STATE	{Set $tag = True$}
\ENDIF
\ENDWHILE
\STATE	{Return $\boldsymbol{\theta}^* = \arg \min_{\boldsymbol{\theta} \in \mc{P}} \mc{L}(\bs{\theta})$}
\end{algorithmic}
\end{algorithm}
%------------------------------------------------------------------

\subsubsection{Mate Selection and Recombination}

Prior to the instance recombination, the {\es} algorithm adopts a mating selection step to pick individuals from the population to become the new parents. Depending on whether the function evaluation is involved in the mate selection or not, the existing mate selection strategies can be divided into two categories:
\begin{itemize}
\item \textbf{Fitness-based Mate Selection}: Such a mate selection approach utilizes the fitness ranking of the parent individuals to pick the good ones for recombination. The global environment selection step (i.e., function call ``select\_$\mu$\_best()'' in Algorithm~\ref{alg:es_algorithm}) can be omitted if this mate selection strategy is adopted.
\item \textbf{Fitness-independent Mate Selection}: This mate selection approach picks the parent individuals doesn't depend on the fitness values of the individuals, which can be either deterministic or stochastic. For such a mate selection strategy, the global environment selection step (i.e., function call ``select\_$\mu$\_best()'' in Algorithm~\ref{alg:es_algorithm}) will be necessary and required.
\end{itemize}

Based on the selected individuals, {\es} will call the ``recombine()'' function to combines information from several parents (the parent individual number is usually denoted by the parameter $\rho$) to generate a single new offspring. There also exist different types of recombination operators in the {\es} algorithm, and several important ones are introduced as follows respectively:
\begin{itemize}
\item \textbf{Discrete Recombination}: Such a recombination operator is also called the uniform crossover in {\ga}, which randomly pick a value from the parents' corresponding variable vector for each of the variable entries. Formally, given the $\rho$ parents available for recombination, for each entry $\bs{\theta}_i'(j), \forall j \in \{1, 2, \cdots, d_{\theta}\}$ in the offspring variable vector, the discrete recombination approach will select on individual from those $\rho$ parents (e.g., $\bs{\theta}_k$) and fill in the entry with values from $\bs{\theta}_k$ (i.e., set $\bs{\theta}_i'(j) = \bs{\theta}_k(j)$).

\item \textbf{Intermediate Recombination}: This recombination operator computes the average value of each variable of the selected $\rho$ parent individuals as the corresponding variable vector for the newly generated offspring. Formally, we can represent the offspring variable vector as $\bs{\theta}_i = \frac{1}{\rho} \sum \bs{\theta}_k$, where the summation term will sums all the variable vectors of the $\rho$ selected parent individuals.

\item \textbf{Weighted Recombination}: The weighted recombination operator is a generalization of the intermediate recombination, which consider the variable vectors from all the $\rho$ parent individuals (usually $\rho = \mu$) via a weighted sum. The weight values are computed based on the fitness ranking of the individuals, where the good individuals will get a weight no less than that of the inferior ones. 
\end{itemize}

Similar to the crossover operator in {\ga}, the recombination operator in {\es} will create variations in the population, which allows the algorithm to explore different regions of the search space. The discrete recombination works quite similar to the crossover operator, which adjust the search regions located at the vertices of the search region; while the intermediate recombination and weighted recombination allow the algorithm to search along the edges of the hyper-rectangle of the search region instead.

\subsubsection{Mutation and Parameter Control}

The mutation operation will introduce some ``small'' variation to the variable vectors, which allows the {\es} to jump to other search regions for further explorations. {\es} introduces the perturbation vector from a multivariate normal distribution, e.g., $\mb{\epsilon} \sim \mc{N}(\mb{0}, \mb{C})$, with zero mean and covariance matrix $\mb{C} \in \mathbbm{R}^{d_{\theta} \times d_{\theta}}$. Formally, as shown in Algorithm~\ref{alg:es_algorithm}, given the variable vector $\bs{\theta}'$ obtained from the recombination, we can represent its corresponding vector after mutation as $\bs{\theta}'' = \bs{\theta}' + \mb{\epsilon}$, which can be treated as a vector draw from distribution $\bs{\theta}' + \mc{N}(\mb{0}, \mb{C})$ (or the equivalent distribution $\mc{N}(\bs{\theta}', \mb{C}^{\frac{1}{2}} \mc{N}(\mb{0}, \mb{I}))$).

%------------------------
\begin{figure}[t]
    \centering
    \includegraphics[width=0.8\textwidth]{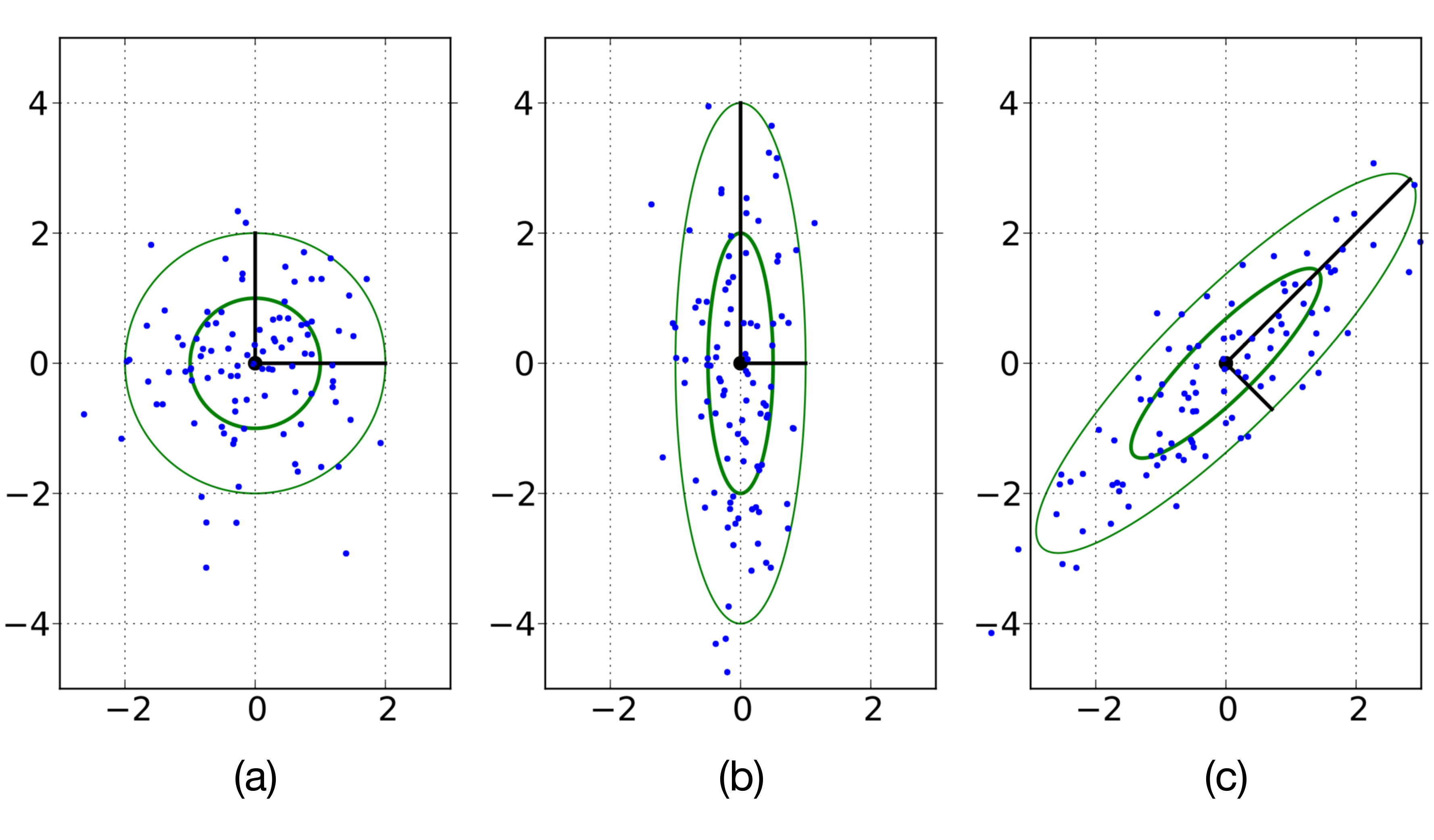}
    \caption{Examples of {\es} Mutation Distribution.}
    \label{fig:chap_derivative_free_es_distribution}
\end{figure}
%------------------------

Therefore, the covariance matrix $\mb{C}$ determines the mutation step in the {\es} algorithm, and the selection of different type of matrix $\mb{C}$ will lead to different types of mutations in {\es}. In Figure~\ref{fig:chap_derivative_free_es_distribution}, we show several examples of the mutation distribution plots with different covariance matrices $\mb{C}$.
\begin{itemize}
\item If the different dimensions of the mutation distribution are independent but with common variance, i.e., $\mb{C} = c \cdot \mb{I}$ (which denotes $\mb{C}$ is a diagonal matrix with constant $c$ on its diagonal), the corresponding distribution $\mb{C}^{\frac{1}{2}} \mc{N}(\mb{0}, \mb{I})$ will be a re-scaled normal Gaussian distribution and its distribution plot will be in a spherical shape (as illustrated in plot (a) of Figure~\ref{fig:chap_derivative_free_es_distribution}).
\item If the different dimensions of the mutation distribution are independent but with different variances, i.e., $\mb{C} = diag(\mb{\sigma}^2)$ (here $diag(\mb{\sigma}^2)$ denotes a diagonal matrix with $\mb{\sigma}^2$ on its diagonal), the distribution region of distribution $\mb{C}^{\frac{1}{2}} \mc{N}(\mb{0}, \mb{I})$ will be in a ellipsoid shape with principal axes parallel to the coordinate axes (as illustrated in plot (b) of Figure~\ref{fig:chap_derivative_free_es_distribution}).
\item If the covariance matrix is positive definite (i.e., $\mb{x}^\top\mb{C}\mb{x} >0, \forall \mb{x} \in \mathbbm{R}^{d_{\theta}}$), which denotes a general case covering the previous two special case as well, the corresponding distribution plot will be in a spherical shape with principal axes pointing to any directions (as illustrated in plot (c) of Figure~\ref{fig:chap_derivative_free_es_distribution}).
\end{itemize}
In the {\es} algorithm, controlling the parameter $\mb{C}$ is the key to design the evolution strategy. Several different parameter control approaches have been introduced, including the {1/5 success rule for parameter control} \cite{rules}, {self-adaption} \cite{adaption}, {derandomized self-adaption} \cite{self-adaption}, \textit{cumulative step-size control} (CSA) \cite{csa, csa2}, \textit{covariance matrix adaption} (CMA) \cite{cmaes} and \textit{natural evolution strategies} (NES) \cite{nes}. In the following part, we will introduce one of them, i.e., CMA, and the {\es} algorithm with CMA based parameter control is also called the {\cmaes}.

\subsubsection{CMA-ES}

%------------------------------------------------------------------
%\setlength{\textfloatsep}{0pt}
\begin{algorithm}[t]
\small
\caption{CMA-ES Algorithm}
\label{alg:cmaes_algorithm}
\begin{algorithmic}[1]
	\REQUIRE Variable search space $\Theta$; Variable dimension $d_{\theta}$; Evolution parameters $\rho$, $\mu$ and $\lambda$
\ENSURE  Model Parameter $\boldsymbol{\theta}$
\STATE	{Set parameters to be used in the algorithm}
\STATE	{Set evolution path vectors $\mb{p}^{(0)} = \mb{0}$ and $\mb{q}^{(0)} = \mb{0}$, covariance matrix $\mb{C}^{(0)} = \mb{I}$ and $g = 0$}
\STATE	{Initialize mean vector $\mb{m}^{(0)} \in \mathbbm{R}^{d_{\theta}}$ with step-size $\sigma^{(0)} \in \mathbbm{R}$}
\STATE	{Initialize convergence $tag = False$}
\WHILE	{$tag = False$}
\STATE	{Initialize population set $\mc{P} = \{\}$}
\FOR	{$i \in \{1, 2, \cdots, \lambda\}$}
\STATE	{// ** \textbf{Sample population individuals:} **}
\STATE	{$\bs{\theta}^{(g+1)}_i \sim \mb{m}^{(g)} + \sigma^{(g)} \mc{N}(\mb{0}, \mb{C}^{(g)})$}
\STATE	{$\mc{P} = \mc{P} \cup \left\{( \bs{\theta}_i, \mc{L}(\bs{\theta}_i) )\right\}$}
\ENDFOR

\STATE	{\ }
\STATE	{// ** \textbf{Update Mean Vector:} **}
\STATE	{$\mb{m}^{(g+1)} = \mb{m}^{(g)} + c_m \cdot \sum_{i = 1}^{\mu} w_i \cdot (\bs{\theta}^{(g+1)}_i - \mb{m}^{(g)})$}

\STATE	{\ }
\STATE	{// ** \textbf{Update covariance matrix:} ** }
\STATE	{$\mb{p}^{(g+1)} = (1-c_c) \mb{p}^{(g)} + \sqrt{c_c (2-c_c) \mu_{eff}} \frac{\mb{m}^{(g+1)} - \mb{m}^{(g)}}{\sigma^{(g)}}$}
\STATE	{$\mb{C}^{(g+1)} = (1 - c_1 - c_{\mu} \sum_{i=1}^{\lambda} w_i) \mb{C}^{(g)} + c_1 \mb{p}^{(g+1)} {\mb{p}^{(g+1)}}^\top + c_{\mu} \sum_{i=1}^{\lambda} w_i (\bs{\theta}^{(g+1)}_i - \mb{m}^{(g+1)})(\bs{\theta}^{(g+1)}_i - \mb{m}^{(g+1)})^\top$}

\STATE	{\ }
\STATE	{// ** \textbf{Update step-size:} ** }
\STATE	{$\mb{q}^{(g+1)} = (1-c_{\sigma}) \cdot \mb{q}^{(g)} + \sqrt{c_{\sigma} (2-c_{\sigma})\mu_{eff}} \cdot ({\mb{C}^{(g)}})^{-\frac{1}{2}} \frac{\mb{m}^{(g+1)} - \mb{m}^{(g)}}{\sigma^{(g)}}$}
\STATE	{$\sigma^{(g+1)} = \sigma^{(g)} \exp \left( \frac{c_{\sigma}}{d_{\sigma}} \left( \frac{\left\| \mb{q}^{(g+1)}\right\|}{\mathsf{E} \left\| \mc{N}(\mb{0}, \mb{I}) \right\|} -1\right) \right)$}

\STATE	{\ }
\IF		{Convergence condition holds}
\STATE	{Set $tag = True$}
\ENDIF
\STATE	{$g = g + 1$}
\ENDWHILE
\STATE	{Return $\boldsymbol{\theta}^* = \arg \min_{\boldsymbol{\theta} \in \mc{P}} \mc{L}(\bs{\theta})$}
\end{algorithmic}
\end{algorithm}
%------------------------------------------------------------------

In this part, we will take the {\cmaes} algorithm \cite{cmaes} as an example to introduce a concrete implementation of {\es}. The key necessary steps involved in {\cmaes} include:
\begin{itemize}
\item \textbf{Mutation}: In {\cmaes}, a population of new search points will be generated by sampling a multivariate normal distribution. Formally, let $\mb{m}^{(g)}$, $\sigma^{(g)}$ and $\mb{C}^{(g)}$ denote the updated mean vector, step size and covariance matrix from the $g_{th}$ generation. Based on them, we can generate the variable individuals in the $(g+1)_{th}$ generation with
\begin{equation}
\bs{\theta}^{(g+1)}_i \sim \mb{m}^{(g)} + \sigma^{(g)} \mc{N}(\mb{0}, \mb{C}^{(g)}), \forall i \in \{1, 2, \cdots, \lambda\}.
\end{equation}
Therefore, we can represent the obtained population for generation $g+1$ as a set $\mc{P} = \{\bs{\theta}^{(g+1)}_1, \bs{\theta}^{(g+1)}_2, \cdots, \bs{\theta}^{(g+1)}_{\lambda}\}$.

\item \textbf{Selection and Recombination}: {\cmaes} adopts the weighted recombination to compute the algorithm new mean for generation $g+1$, which can be denoted as
\begin{equation}
\mb{m}^{(g+1)} = \mb{m}^{(g)} + c_m \cdot \sum_{i = 1}^{\mu} w_i \cdot (\bs{\theta}^{(g+1)}_i - \mb{m}^{(g)}),
\end{equation}
where $c_m \le 1$ denotes the learning rate in {\cmaes}, term $w_i$ denotes the weight of individual $\bs{\theta}_i^{(g+1)}$ and we have $\sum_{i = 1}^{\mu} w_i = 1$. Here, individuals in the population set $\mc{P}$ will be sorted according to their evaluations of the objective function, and the top $\mu$ individuals are selected from the population set $\mc{P}$. In other words, the mate selection strategy used in {\cmaes} is fitness dependent. Different ways can be used to define the weight terms $\{w_1, w_2, \cdots, w_{\mu}\}$, and setting $w_i = \frac{1}{\mu}, \forall i \in \{1, 2, \cdots, \mu\}$ will reduce the recombination to the intermediate recombination.

\item \textbf{Covariance Matrix Adaption}: {\cmaes} updates the covariance matrix iteratively, and the updating equation considers information from three different perspectives: (1) current covariance matrix, (2) covariance computed with the newly mutated individuals, and (3) covariance computed with the evolution path. Formally, the covariance matrix updating equation can be represented as follows:
\begin{equation}
\mb{C}^{(g+1)} = (1 - c_1 - c_{\mu} \sum_{i=1}^{\lambda} w_i) \mb{C}^{(g)} + c_1 \mb{p}^{(g+1)} {\mb{p}^{(g+1)}}^\top + c_{\mu} \sum_{i=1}^{\lambda} w_i (\bs{\theta}^{(g+1)}_i - \mb{m}^{(g+1)})(\bs{\theta}^{(g+1)}_i - \mb{m}^{(g+1)})^\top,
\end{equation}
where these three terms denote the covariance matrices mentioned above, and $c_1$, $c_{\mu}$ represent the weights of the last two terms. Vector $\mb{p}^{(g+1)}$ denote the evolution path vector of the $(g+1)_{th}$ generation, and it can be denoted as
\begin{align}
\mb{p}^{(g+1)} = (1-c_c) \mb{p}^{(g)} + \sqrt{c_c (2-c_c) \mu_{eff}} \frac{\mb{m}^{(g+1)} - \mb{m}^{(g)}}{\sigma^{(g)}},
\end{align}
where $\mu_{eff} = (\sum_{i=1}^{\mu}w_i^2)^{-1}$ is the variance effective selection mass and $c_c \le 1$ is a weight term. The factor $\sqrt{c_c (2-c_c) \mu_{eff}}$ is a normalization constant for $\mb{p}$. For $c_c =1$ and $\mu_{eff} = 1$, the factor reduces to $1$ and $\mb{p}^{(g+1)}$ reduces to $\mb{p}^{(g+1)} = \frac{\mb{m}^{(g+1)} - \mb{m}^{(g)}}{\sigma^{(g)}}$.

\item \textbf{Step-Size Control}: In addition to the covariance matrix adaption, {\cmaes} will also update the step-size which controls the variation scale in the mutation. Formally, the updating equation of the step-size can be denoted with the following equation:
\begin{equation}
\sigma^{(g+1)} = \sigma^{(g)} \exp \left( \frac{c_{\sigma}}{d_{\sigma}} \left( \frac{\left\| \mb{q}^{(g+1)}\right\|}{\mathsf{E} \left\| \mc{N}(\mb{0}, \mb{I}) \right\|} -1\right) \right).
\end{equation}
In the above equation, $d_{\sigma} \approx 1$ is the damping parameter, and vector $\mb{q}^{(g+1)}$ denote the conjugate evolution path vector in generation $g+1$, which can be denoted as
\begin{equation}
\mb{q}^{(g+1)} = (1-c_{\sigma}) \cdot \mb{q}^{(g)} + \sqrt{c_{\sigma} (2-c_{\sigma})\mu_{eff}} \cdot ({\mb{C}^{(g)}})^{-\frac{1}{2}} \frac{\mb{m}^{(g+1)} - \mb{m}^{(g)}}{\sigma^{(g)}},
\end{equation}
where $c_{\sigma} < 1$ is a weight term.
\end{itemize}

Based on these steps introduced above, we can provide the pseudo-code of the {\cmaes} algorithm in Algorithm~\ref{alg:cmaes_algorithm}.

%----------------------------------------------------------------------------------------------
%----------------------------------------------------------------------------------------------
%----------------------------------------------------------------------------------------------

\section{Random Search Algorithms}

In this part, we will introduce several other derivative-free optimization algorithms based on generic random search, which don't belong to the above three categories of algorithms that we have introduced before in this paper and in \cite{zhang2019derivativefree_part1}. Many of the algorithms introduced above actually may also belong to the random search algorithm category, e.g., {\ga} and {\es}. Random search algorithms are useful for many ill-structured global optimization problems with continuous and/or discrete variables. The random search algorithms to be introduced here include {Hill Climbing} and {Simulated Annealing}.

% \cite{}, {Cross-Entropy (CE)} \cite{}, {Data-based Online Nonlinear Extremum-seeker (DONE)} \cite{} and {Ant Colony Optimization} \cite{}.

%----------------------------------------------------------------------------------------------

\subsection{Hill Climbing}

The hill climbing algorithm to be introduced here has a close relation with the gradient descent algorithms we introduced in the previous tutorial article \cite{zhang2019gradient}, which all search for the solutions via an iterative search to maximize/minimize the function evaluation at the current state. Meanwhile, hill climbing is also very different from gradient descent, since it doesn't require any derivative computation in the optimization process. In hill climbing, starting at the base of a hill, we walk upwards until we reach the top of the hill. In other words, hill climbing starts with initial state and keeps improving the solution until reaching its (local) optimum.

%------------------------------------------------------------------
%\setlength{\textfloatsep}{0pt}
\begin{algorithm}[t]
\small
\caption{Hill Climbing Algorithm}
\label{alg:hill_climbing}
\begin{algorithmic}[1]
	\REQUIRE Variable search space $\Theta$;
\ENSURE  Model Parameter $\boldsymbol{\theta}$
\STATE	{Initialize a random variable vector $\bs{\theta} \in \Theta$}
\STATE	{Initialize the stop $tag = False$}
\WHILE	{$tag = False$}
\STATE	{Compute the neighbor set $\Gamma(\bs{\theta})$ of $\bs{\theta}$}
\STATE	{Evaluate the function and compute $\{\mc{L}(\bs{\theta}_i)\}_{\bs{\theta}_i \in \Gamma(\bs{\theta})}$}
\STATE	{Set $\bs{\theta}'$ to be the optimal neighbor in $\Gamma(\bs{\theta})$ with the largest function evaluation value}
\IF		{$\mc{L}(\bs{\theta}') \le \mc{L}(\bs{\theta})$}
\STATE	{Set $tag = True$}
\ELSE
\STATE	{Set $\bs{\theta} = \bs{\theta}'$}
\ENDIF
\ENDWHILE
\STATE	{Return $\bs{\theta}$ as the solution}
\end{algorithmic}
\end{algorithm}
%------------------------------------------------------------------

The pseudo-code of the hill climbing algorithm is provided in Algorithm~\ref{alg:hill_climbing}. The algorithm involves several key steps:
\begin{itemize}

\item \textbf{Initialization}: The hill climbing algorithm starts with a random point in the search space, which can be denoted as $\bs{\theta} \in \Theta$.

\item \textbf{Neighborhood Exploration}: Based on the current state, the hill climbing algorithm iteratively searches the nearby points of $\bs{\theta}$ to determine the moving direction for the next step. Formally, these nearby points define the neighbor set of $\bs{\theta}$, which can be denoted as $\Gamma(\bs{\theta})$. Different ways can be used to define the ``neighbor'' in hill climbing. For instance, given a vector $\bs{\theta}$, we can enumerate all the entries in the vector to add/minus $1$ to define a group of new nearby vectors as the neighbors, which can be denoted as $\{[\bs{\theta}(1), \bs{\theta}(2), \cdots, \bs{\theta}(i) \pm 1, \cdots, \bs{\theta}(d_{\theta})  ]\}_{i \in \{1, 2, \cdots, d_{\theta}\}}$.

\item \textbf{Neighborhood Evaluation}: Hill climbing will evaluate the objective function $\mc{L}(\cdot)$ at these neighbor points, and pick the optimal one with the minimum function evaluation, e.g., $\bs{\theta}'$. 

\item \textbf{Stop Criterion}: If the optimal neighbor variable is better than the current variable, i.e., $\mc{L}(\bs{\theta}') \ge \mc{L}(\bs{\theta})$, the algorithm will moves to that neighbor point. Otherwise, the algorithm will stop and return the current variable as the output.

\end{itemize}

%------------------------
\begin{figure}[t]
    \centering
    \includegraphics[width=0.8\textwidth]{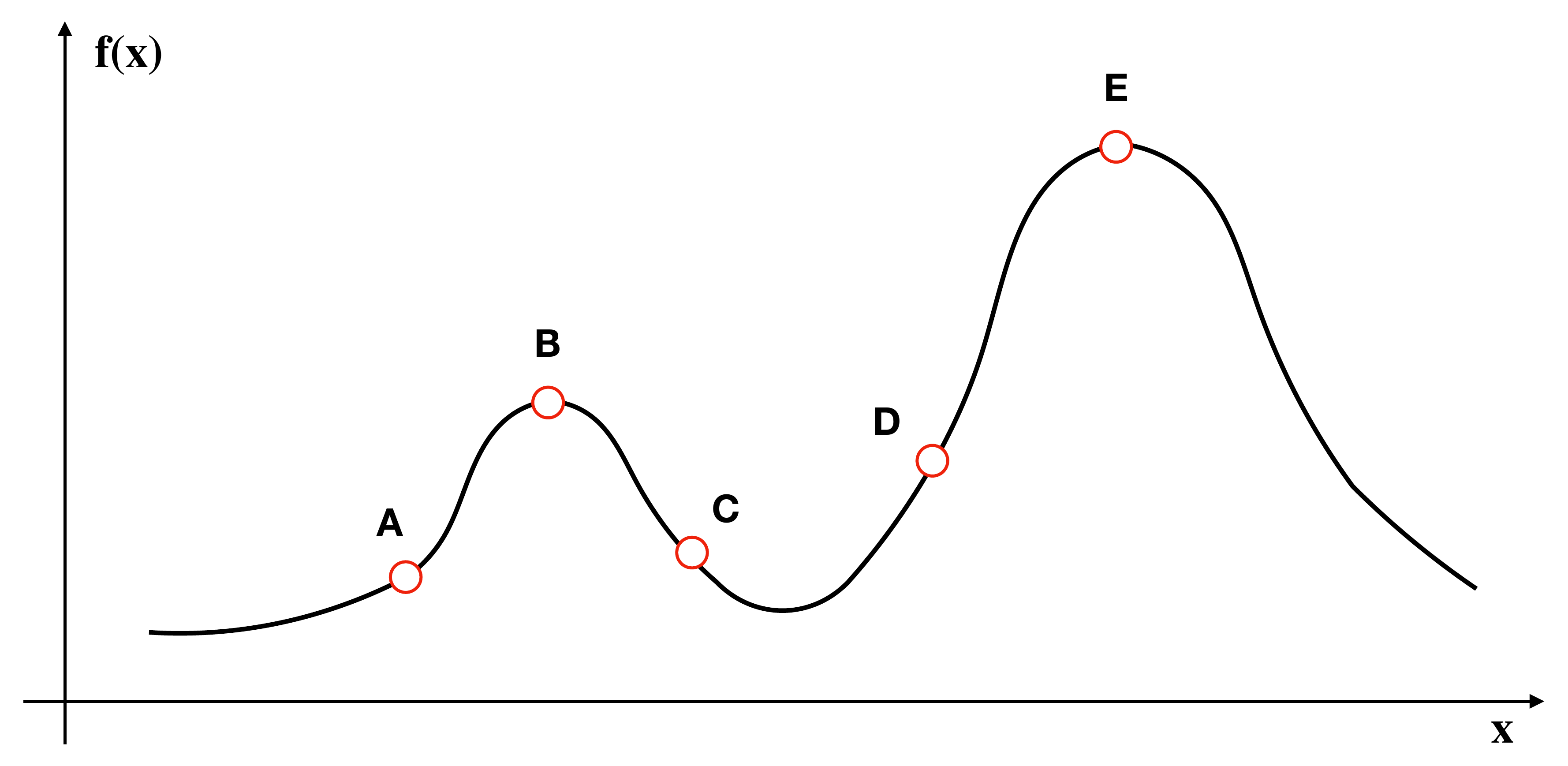}
    \caption{An Example of the Hill Climbing Algorithm.}
    \label{fig:chap_derivative_free_hill_climbing}
\end{figure}
%------------------------

\begin{example}\label{exp:climbing_hill_example}
For instance, in Figure~\ref{fig:chap_derivative_free_hill_climbing}, we provide an example to illustrate how the hill climbing algorithm works in learning the optimal variable to the single-variable function $f(x)$. Let $A$ denote the starting point, where optimization process starts. The hill climbing algorithm will select its neighbor point $B$ as the successor to evaluate the function. We know that $f(B) > f(A)$. The algorithm will move the current state to $B$ and the search process will continue to point $C$. Noticing that $f(C) < f(B)$, the climbing algorithm will stop there and return $B$ as the optimal solution, which is actually a local maximum point of the function.

\end{example}

%----------------------------------------------------------------------------------------------

\subsection{Simulated Annealing}

According to the above description as well as Example~\ref{exp:climbing_hill_example}, we can discover that the hill climbing algorithm is a greedy algorithm and can be short-sighted in the optimization process, especially for the non-convex functions. For instance, in Example~\ref{exp:climbing_hill_example}, the algorithm is only able to identify a local maximum $B$ but fail discover the global optima $E$, which is nearby to $B$ actually. In this part, we will introduce the simulated annealing algorithm, which is a probabilistic technique for approximating the global optimum of a given function. Simulated annealing allows the continuous search in the worse-regions subject to a decreasing probability.

The name, i.e., ``simulated annealing'', and inspiration come from annealing in metallurgy, a technique involving heating and controlled cooling of a material to increase the size of its crystals and reduce their defects. The simulation of annealing can be used to find an approximation of a global minimum for a function with a large number of variables. The simulated annealing algorithm allows the search in the region which is worse than the current state, but the probability for such a kind of worse-region search decreases step by step. This notion of slow cooling implemented in the simulated annealing algorithm is interpreted as a slow decrease in the probability of accepting worse solutions as the solution space is explored. Accepting worse solutions is a fundamental property of meta-heuristics because it allows for a more extensive search for the global optimal solution.

In general, the simulated annealing algorithms work as follows. 
\begin{itemize}
\item The algorithm initializes the search by randomly selecting one point in the search space.
\item At each time step, the algorithm randomly selects a solution close to the current one, measures its quality, and then decides to move to it or to stay with the current solution.
\item During the search, the temperature is progressively decreased from an initial positive value to zero and affects the worse-region exploration probabilities, i.e., the probability of moving to a worse new solution is progressively changed towards zero.
\end{itemize}
 
 %------------------------------------------------------------------
%\setlength{\textfloatsep}{0pt}
\begin{algorithm}[t]
\small
\caption{Simulated Annealing Algorithm}
\label{alg:simulated_annealing_algorithm}
\begin{algorithmic}[1]
	\REQUIRE Variable search space $\Theta$; Temperature decreasing factor $\alpha \in [0, 1]$; Max iteration $I_{max}$
\ENSURE  Model Parameter $\boldsymbol{\theta}$
\STATE	{Initialize a random variable vector $\bs{\theta} \in \Theta$}
\STATE	{Initialize iteration count $c = 0$}
\WHILE	{iteration $c \le I_{max}$}
\STATE	{Compute the neighbor set $\Gamma(\bs{\theta})$ of $\bs{\theta}$}
\STATE	{Select a point $\bs{\theta}'$ from set $\Gamma(\bs{\theta})$}
\IF		{$\mc{L}(\bs{\theta}') \ge \mc{L}(\bs{\theta})$}
\STATE	{Set $\bs{\theta} = \bs{\theta}'$}
\ELSE
\IF		{$\exp\left( \frac{\mc{L}(\bs{\theta}') - \mc{L}(\bs{\theta}) }{T} \right) > random(0,1)$}
\STATE	{Set $\bs{\theta} = \bs{\theta}'$}
\ENDIF
\ENDIF
\STATE	{Update $T = \alpha \cdot T$}
\STATE	{Update $c = c + 1$}
\ENDWHILE
\STATE	{Return $\bs{\theta}$ as the solution}
\end{algorithmic}
\end{algorithm}
%------------------------------------------------------------------

The pseudo-code of the simulated annealing algorithm is provided in Algorithm~\ref{alg:simulated_annealing_algorithm}. According to the pseudo-code, the algorithm accepts the temperature decreasing factor $\alpha \in [0,1]$ and the maximum iteration $I_{max}$ as the input. Via an iterative search, the algorithm will accept a new neighbor point $\bs{\theta}'$ iff (1) $\mc{L}(\bs{\theta}') \ge \mc{L}(\bs{\theta})$; or (2) $\mc{L}(\bs{\theta}') < \mc{L}(\bs{\theta})$ but $random(0,1) < \exp\left( \frac{\mc{L}(\bs{\theta}') - \mc{L}(\bs{\theta}) }{T} \right)$. In other words, the algorithm will explore to $\bs{\theta}'$ if it leads to a better performance, or it will explore a worse region with probability $P(T; \bs{\theta}', \bs{\theta}) = \exp\left( \frac{\mc{L}(\bs{\theta}') - \mc{L}(\bs{\theta}) }{T} \right)$. The probability term $P(T; \bs{\theta}', \bs{\theta})$ depends on both the function evaluations at points $\bs{\theta}'$ and $\bs{\theta}$, as well as the temperature term $T$. Noticing the $T$ term will be updated iteratively with $T = T \times \alpha$ to $0$ as the algorithm continues, which will lower down the probability value $P(T)$ for worse-region exploration steadily.

\begin{example}
For instance, we can still use the example shown in Figure~\ref{fig:chap_derivative_free_hill_climbing}, where $A$ is the starting point for search. The hill climbing algorithm will stop at $B$ since further exploration will lead to worse results. However, the simulated annealing algorithm allows us to take the risk in further searching the region after $C$ with a certain probabilities, e.g., $D$ and $E$, even though $C$ and $D$ are both worse than $B$. It will allow the algorithm to reach the global optimum $E$ at the final.
\end{example}

\section{Summary}

In this paper, as a follow-up of \cite{zhang2019derivativefree_part1}, we have introduce several other categories of derivative-free optimization algorithms, including both the population based algorithms and several other random search based algorithms. Many of these introduced algorithms can be potentially applied to learn the deep neural network models. This tutorial article will also be updated accordingly as we observe more new developments on this topic in the near future.

\newpage

\vskip 0.2in
\bibliographystyle{plain}
\bibliography{reference}

\end{document}